\documentclass[journal]{IEEEtran}
\usepackage{amsmath,graphicx,amssymb}
\usepackage{amsfonts,bm}
\usepackage{pstricks,pst-node}
\usepackage{subfigure}

\title{Optimal Convergence Rate in Feed Forward Neural Networks using HJB Equation}
\author{Vipul Arora, Laxmidhar Behera and Ajay Pratap Yadav%
\thanks{The authors are with the Department of Electrical Engineering, Indian Institute of Technology, Kanpur - 208016 (India). E-mail: vipular.iitk@gmail.com, lbehera@iitk.ac.in, ajaypratapyadav@gmail.com}
}

\begin{document}
\maketitle
\begin{abstract}

A control theoretic approach is presented in this paper for both batch and instantaneous updates of weights in feed-forward neural networks. 
The popular Hamilton-Jacobi-Bellman (HJB) equation has been used to generate an optimal weight update law. The remarkable contribution in this paper is 
that closed form solutions for both optimal cost and weight update can be achieved for any feed-forward network using HJB equation in a simple yet 
elegant manner. The proposed approach has been compared with some of the existing best performing learning algorithms. It is found as expected that the 
proposed approach is faster in convergence in terms of computational time.  
Some of the benchmark 
test data such as 8-bit parity, breast cancer and credit approval, as well as 2D Gabor function have been used to validate our claims. The paper also discusses issues related to global optimization. The limitations of popular deterministic weight update laws are critiqued and the possibility of global optimization using HJB formulation is discussed.
It is hoped that the proposed algorithm will bring in a lot of interest in researchers working in developing fast learning algorithms and global optimization.

\end{abstract}

\section{Introduction}
 
\IEEEPARstart{S}{ince} the advent of the popular back propagation (BP) algorithm \cite{Rumelhart89}, issues concerning global convergence as well as fast convergence have caught the attention of researchers. One of the very first works by Hagan and Menhaj \cite{Hagan1994} makes use of the Levenberg-Marquardt (LM) algorithm for training weights of multi-layered networks in batch mode. Gradient descent scheme that is at the heart of BP  uses only first derivative. Newton's method enhances the performance by making use of the second derivative which is computationally intensive. Gauss Newton's method approximates the second derivative with the help of first derivatives and hence, provides a simpler way of optimization. The Levenberg-Marquardt (LM) algorithm \cite{Hagan1994,Lera2002} modifies the Gauss Newton's method to improve the convergence rate, by interpolating between the gradient descent and the Gauss-Newton method. 
Faster convergence has been reported using extended Kalman Filtering (EKF) approach \cite{Iiguni1992} and recursive least square approach \cite{Bilski1998}. 
Cong and Liang \cite{Cong2009} derive an adaptive learning rate for a particular structure of neural networks using resilient BP algorithm with an aim of stability in uncertain environment.
Man {\sl et al.} \cite{Man2006} use BP with Lyapunov stability theory, while
Mohseni and Tan \cite{Mohseni2012} use BP with variable structure system for robust and fast convergence.
Using Lyapunov function and stability principle, Behera {\sl et al.} \cite{Behera2006} propose a scheme for adaptive learning rate associated with BP algorithm. The above methods ensure fast convergence but can not guarantee global convergence.


\subsection{Global Optimization in FFNN}


Let us consider modulo-2 function $f:\mathbb{R}^2\rightarrow\mathbb{R}$. 
Here, 9 patterns are formed using $x_1\in\{0,1,2\}$ and $x_2\in\{0,1,2\}$.
The desired output is given by
\begin{align}\label{eq:mod2}
y = f(x_1,x_2) = 
\begin{cases} 1 & \text{if } (x_1+x_2) \text{ is odd } \\
  0 & \text{if } (x_1+x_2) \text{ is even }
\end{cases}
\end{align}

The patterns formed by this function are shown in Table~\ref{tab:mod2}.
\begin{table}[h]
\centering
\caption{Patterns formed by Modulo-2 function \label{tab:mod2}}
  \begin{tabular}{|cc|c|}\hline
    $x_1$ & $x_2$ & y \\ [0.5ex] \hline
    0 & 0 & 0 \\
    0 & 1 & 1 \\
    0 & 2 & 0 \\ [0.5ex] \hline
  \end{tabular}
{\hskip 2mm}
\begin{tabular}{|cc|c|}\hline
  $x_1$ & $x_2$ & y \\ [0.5ex] \hline
  1 & 0 & 1 \\
  1 & 1 & 0 \\
  1 & 2 & 1 \\[0.5ex] \hline
\end{tabular}
{\hskip 2mm}
\begin{tabular}{|cc|c|}\hline
  $x_1$ & $x_2$ & y \\ [0.5ex] \hline
  2 & 0 & 0 \\
  2 & 1 & 1 \\
  2 & 2 & 0 \\[0.5ex] \hline
\end{tabular}
\end{table}

\begin{figure}
\begin{center}
  \scalebox{0.5}{\input{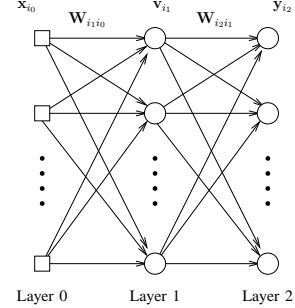}}
  \caption{Schematic of a two-layer FFNN.\label{fig:FFNN}}
\end{center}
\end{figure}
We consider a two-layer FFNN, with $N_{i_0}$-$N_{i_1}$-$N_{i_2}$ structure, where $N_{i_0},N_{i_1},N_{i_2}$ stand for the number of neurons in the input, hidden and output layers, respectively. Fig. \ref{fig:FFNN} shows a block schematic of the same.
The non-linearity is introduced by using a sigmoidal activation function at each neuron.
\begin{align}
h_{i_1} &= \sum_{i_0}w_{i_1i_0}x_{i_0} \\
v_{i_1} &= \frac{1}{1+e^{-h_{i_1}}} \label{eq:NN1}\\
h_{i_2} &= \sum_{i_1}w_{i_2i_1}v_{i_1} \\
y_{i_2} &= \frac{1}{1+e^{-h_{i_2}}} \label{eq:NN2}
\end{align}

\begin{figure}
\subfigure[]{\includegraphics[width = 0.48\linewidth, height = 0.45\linewidth]{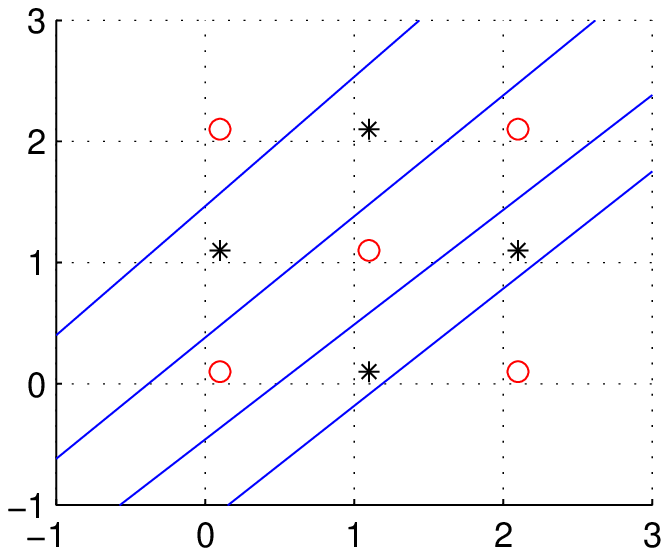}}
\subfigure[]{\includegraphics[width = 0.48\linewidth, height = 0.45\linewidth]{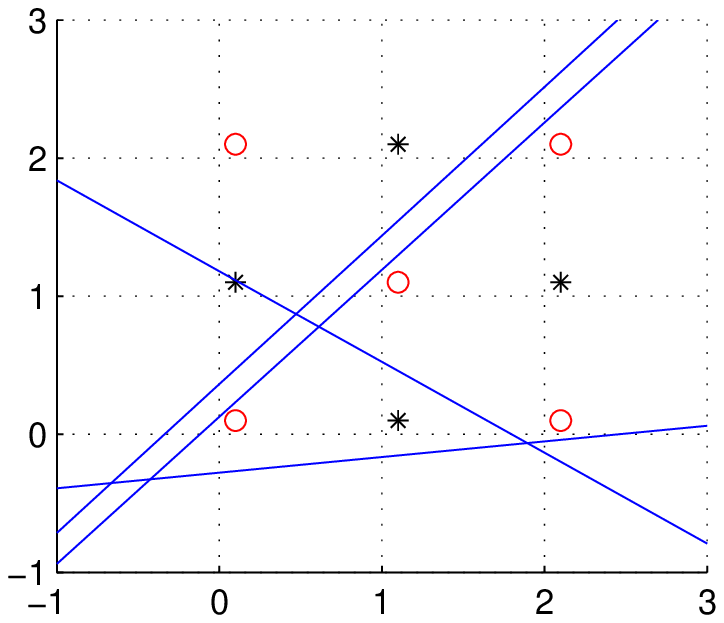}}
\caption{Modulo-2 function: the two axes carry the two inputs; red circles and black stars represent output as 0 and 1, respectively. Blue lines show the separators of the layer 1 of a 3-4-1 FFNN. (a) Global minimum reached; (b) Weights stuck in a local minimum. \label{fig:mod2}}
\end{figure}

Each neuron acts as a linear separator which divides its input space into two regions. Its output changes smoothly, from 0 in one region to 1 in the other, along the sigmoidal function. For the modulo-2 function, Fig. \ref{fig:mod2}(a) shows one possible configuration of separators in the layer-1. The output ($\in\mathbb{R}^4$) of layer-1 for this configuration is linearly separable. This shows that the modulo-2 function can be modeled with the help of a 3-4-1 FFNN, where a bias (+1) is the third input.
Also, since the pattern space is symmetric about the line $x_1=1$, we can get another configuration of separators reflected about $x_1=1$. This shows that multiple global minima may be possible for a problem. However, local minima also exist for this example, such as the one shown in Fig. \ref{fig:mod2}(b).

The {\sl success score} is defined as the number of trials which successfully converge to zero output error over the total number of trials.
\begin{align}\label{eq:successscore}
\text{Success score} = \frac{\text{No. of successfully converging trials}}{\text{Total no. of trials}}
\end{align}
If the success score is 100\% then it implies that the network converges to the global minimum always.

For training an FFNN, several trials are run, each initializing from a different set of weights. For implementation, an offset of 0.1 is added to all the inputs, and in the output, 0 and 1 are replaced by 0.1 and 0.9, respectively.
For a 3-4-1 FFNN trained using BP algorithm in batch mode, the success score is observed to be around 25$\%$. This means 75$\%$ times the network is stuck in a local minimum. On increasing the number of hidden neurons to 6, the success score rises to around 80$\%$, and further to 90$\%$ with 10 hidden neurons. The success score reaches close to 100$\%$ with 15 hidden neurons. 

This simple experiment shows that the mapping of the modulo-2 function using 3-4-1 FFNN generates both local and global minima in error surface. However by increasing the number of hidden neurons to 15 or beyond, the error surface ceases to have local minima as the above experiment shows. Instead if we consider a XOR function, then the score becomes always 100$\%$ irrespective of the network architecture as there is no local minimum in the error-surface. This experiment amply suggests that the type of the function and  the network architecture together determine if the error surface will have local minimum. This modulo-2 function has been used in this paper as a benchmark to study global convergence in a FFNN.
    


\subsection{Previous Approaches}
The approaches for optimization can broadly be classified into two categories: deterministic and probabilistic.

The deterministic methods aim at finding the minimum of the objective function by evaluating the function iteratively in steps, where each step is determined using the derivatives of the function. A simple local search method is gradient descent based search which steps in the direction of gradient of the function. For FFNN, back propagation (BP) training is based on the gradient descent scheme.
In order to find a global minimum, a popular method is to move from the current local minimum to a lower minimum by transforming the objective function. 
Ng {\sl et al.} \cite{Ng2010} discuss various auxiliary functions which are designed to transport the algorithm from current minimum to basins of better minima, using gradient descent. 
For training of FFNN's, Shang and Wah \cite{Shang1996} use a trajectory based method that depends on heuristic functions to search the minima.
Toh \cite{Toh2003} uses deterministic line search over the monotonic transformation of error function to progress to lower minima.
Gan and Li \cite{Gan2014} use LM algorithm for non-linear least square optimization. 

The probabilistic methods, on the other hand, may either proceed in steps, with step size defined stochastically, or else choose initial points for deterministic local searches.
Simulated annealing associates a time varying probability distribution function with the parameter search space, such that the parameter configurations with lower values of objective function have more probability. The parameters changes from one configuration to another based on stochastic jumps until they reach the configuration with maximum probability.
Another popular approach uses genetic algorithms, which use heuristically rules inspired from the biological evolution models.
For training FFNN's, Boese and Kahng \cite{Boese1993} use simulated annealing, while Tsai \emph{et al.} \cite{Tsai2006} and Roh \emph{et al.} \cite{Roh2007} use genetic algorithms.
Sexton \emph{et al.} \cite{Sexton1999} compare the genetic algorithms and simulated annealing based approaches.
Lidermir \emph{et al.} \cite{Ludermir2006} combine simulated annealing and tabu search with gradient descent so as to optimize the FFNN weights and architecture.
Delgado \emph{et al.} \cite{Delgado2008} use hybrid algorithms for evolutionary training of recurrent NN's.

In this paper, we want a theoretical investigation if there exists a possibility of finding a deterministic weight update law that can always ensure global minimum, starting from any random initialization of weights.

\subsection{Contribution and organization of the paper}
Researchers have tried to solve global optimization in FFNN using random weight search and such approaches have remained heuristic at best. The deterministic approaches to global optimization \cite{Toh2003} is highly dependent on search direction whose computation is again a heuristic process. We are surprised to find that nobody has ever thought of using Hamilton-Jacobi-Bellman (HJB) equation for solving this problem. The HJB equation comes from dynamic programming \cite{BeheraKar2009}, which is a popular approach for optimal control of dynamical systems \cite{Ferrari2008,Wei2014}.

In this paper, the weight update law has been converted into a control problem and dynamic optimization has been used to derive the update law. 
The derivation of the HJB based weight update law is surprisingly simple and straightforward. The closed form solution for the optimal cost and optimal weight update law have simple structures. The proposed approach has been compared with some of the existing best performing learning algorithms and is found to be faster in convergence in terms of computational time. 
In this paper we are investigating if HJB based weight update law can address the following two issues: 

\begin{enumerate}
\item Can the local minima in 3-4-1 network architecture for modulo 2 function be avoided? 
\item Can the algorithm converge to the local minimum in optimum speed for any problem?
\end{enumerate}

The rest of the paper has been organized as follows. Sec.~\ref{sec:offlineHJB} represents the supervised training of FFNN as a control problem and derives optimal weight update law using HJB equations. Sec.~\ref{sec:onlineHJB} derives HJB based weight update law in instantaneous mode, with a special case of single output neural network. Simulation results are presented in Sec.~\ref{sec:Results}. 
A detailed discussion on convergence behavior has been made in Sec.~\ref{sec:globalMin}. Concluding remarks are provided in Sec.~\ref{sec:Concl}.  

\section{HJB based Offline Learning of FFNN}\label{sec:offlineHJB}
The output $\mathbf{y}_p\in\mathbb{R}^{N_o}$ for a given input pattern $\mathbf{x}_p\in\mathbb{R}^{N_i}$ can be written as,
\begin{equation}
\mathbf{y}_p = \mathbf{f}(\mathbf{\widehat{w}},\mathbf{x}_p)
\end{equation}
Here, $\mathbf{\widehat{w}}\in \mathbb{R}^{N_w}$ is the vector of weight parameters involved in the FFNN to be trained, with total $N_w$ weight parameters.
The derivative of $\mathbf{y}$ w.r.t. time $t$ is
\begin{align}
\mathbf{\dot{y}}_p = \frac{\partial{\mathbf{f}(\mathbf{\widehat{w}},\mathbf{x}_p)}}{\partial{\mathbf{\widehat{w}}}} \mathbf{\dot{\widehat{w}}} = \mathbf{J}_p \mathbf{\dot{\widehat{w}}}
\end{align}
where, $\mathbf{J}_p= \frac{\partial{\mathbf f(\mathbf{\widehat{w}},\mathbf x_p)}}{\partial{\mathbf{\widehat{w}}}}$ is the Jacobian matrix, whose elements are
${J}_{p,ij}={\partial{{y}_{p,i}}} / {\partial{{w}_j}}.$
The desired output is given by
$\mathbf{y}_p^d = \mathbf{f}(\mathbf{w},\mathbf{x}_p)$
and its derivative with respect to time is
\begin{equation}
\mathbf{\dot{y}}_p^d = \mathbf{J}_p\mathbf{\dot{w}} = 0.
\end{equation}
The estimation error is
$\mathbf{e}_p = \mathbf{y}^d_p-\mathbf{y}_p$
and its derivative w.r.t. time $t$ is
\begin{align}
\mathbf{\dot{e}}_p = \mathbf{\dot{y}}_p^d -\mathbf{\dot{y}}_p = -\mathbf{J}_p \dot{\mathbf{\widehat{w}}} , \text{ as } \mathbf{\dot{y}}_p^d=0
\end{align}

The optimization of the neural network weights is formulated as a control problem. 
\begin{align}
\mathbf{\dot{e}}_p = -\mathbf{J}_p \dot{\mathbf{\widehat{w}}} = -\mathbf{J}_p \mathbf{u} \label{eq:FFNNdyn}
\end{align}
The control input updates the weights as $\mathbf{u} = \mathbf{\dot{\widehat{w}}}$.

In the batch mode, all the patterns are learnt simultaneously. 
Hence, the dynamics can be written jointly as
\begin{align} \label{eq:edot}
\mathbf{\dot{e}} & = -\mathbf{J} \mathbf{u}
\end{align}
where, $\mathbf{e} = [\mathbf{e}_1^\intercal, \mathbf{e}_2^\intercal, ..., \mathbf{e}_{N_p}^{\intercal}]^\intercal $, $\mathbf{J}=[\mathbf{J}_1^\intercal, \mathbf{J}_2^\intercal, ... , \mathbf{J}_{N_p}^\intercal]^\intercal $.
Hence, $\mathbf{e}$ is an $N_oN_p\times 1$ vector, $\mathbf{J}$ is an $N_oN_p\times N_w$ matrix and $\mathbf{u}$ is an $N_w\times 1$ vector.

\subsection{Optimal Weight Update}
The cost function is defined over time interval $(t,T]$ as
\begin{align} \label{eq:costFnV}
V(\mathbf{e}(t)) &= \int_t^T L(e(\tau),\mathbf{u}(\tau))d\tau\\
\text{where, } 
L(\mathbf{e},\mathbf{u}) &= \frac{1}{2}(\mathbf{e}^\intercal\mathbf{e} + \mathbf{u}^\intercal \mathbf{R}\mathbf{u}) \label{eq:L}
\end{align}
with $\mathbf{R}$ as a constant $N_w\times N_w$ matrix. Here, $t$ signifies the iterations to update $\mathbf{w}$.
Our goal is to find an optimal weight update law $\mathbf{u}(t)$ which minimizes the aforementioned global cost function. 
Hence, we get the following equation which is popularly known as the Hamilton-Jacobi-Bellman (HJB) equation.
\begin{align}
\min_{\mathbf{u}}\{\frac{dV^*}{d\mathbf{e}}\mathbf{\dot{e}}(t) + L(\mathbf{e}(t),\mathbf{u}(t))\} = 0
\end{align}
Putting the expressions for $\mathbf{\dot{e}}(t)$ and $L(\mathbf{e}(t),\mathbf{u}(t))$ from eqs. \eqref{eq:edot} and \eqref{eq:L}, respectively,
\begin{align}\label{eq:HJB}
\min_{\mathbf{u}}\{-\frac{dV^*}{d\mathbf{e}} \mathbf{J} \mathbf{u}(t) + \frac{1}{2}\mathbf{e}(t)^\intercal \mathbf{e}(t) + \frac{1}{2}\mathbf{u}(t)^\intercal \mathbf{R}\mathbf{u}(t) \} = 0
\end{align}
Here, $\frac{dV^*}{d\mathbf{e}}$ is $1\times N_oN_p$ vector.
Differentiating with respect to $\mathbf{u}$, we get the optimal update law as
\begin{align}\label{eq:ustar}
\mathbf{u}^*(t) = \mathbf{R}^{-1} \mathbf{J}^\intercal \left(\frac{dV^*}{d\mathbf{e}}\right)^\intercal
\end{align}
In order to find the expression for $\left(\frac{dV^*}{d\mathbf{e}}\right)$, we put the optimal $\mathbf{u}$ from eq. \eqref{eq:ustar} in eq. \eqref{eq:HJB},
\begin{align}
\mathbf{e}(t)^\intercal \mathbf{e}(t) - \left(\frac{dV^*}{d\mathbf{e}}\right) \mathbf{J} \mathbf{R}^{-1} \mathbf{J}^\intercal \left(\frac{dV^*}{d\mathbf{e}}\right)^\intercal = 0 \label{eq:dVde_HJB}
\end{align}
A proper solution of eq.~\eqref{eq:dVde_HJB} should lead to $\mathbf{J} \mathbf{R}^{-1} \mathbf{J}^\intercal$ to be positive definite.

This is an under-determined system of equations. 
However, the optimal input must stabilize the system. The stability of the system can be analyzed with the help of a Lyapunov function defined as
\begin{equation}
\mathcal{V}(\mathbf{e}) = \frac{1}{2}\mathbf{e}^\intercal \mathbf{e}
\end{equation}
The equilibrium point $\mathbf e=0$ is stable if $\dot{\mathcal{V}}(\mathbf{e})$ is negative definite.
\begin{align}
\dot{\mathcal{V}}(\mathbf{e}(t)) & = \mathbf{e}^\intercal \mathbf{\dot{e}}\\
& = -\mathbf{e}^\intercal \mathbf{J} \mathbf{u}^*(t)\\
& = -\mathbf{e}^\intercal \mathbf{J} \mathbf{R}^{-1} \mathbf{J}^\intercal \left(\frac{dV^*}{d\mathbf{e}}\right)^\intercal
\end{align}


If one selects the following form of ${dV^*}/{d\mathbf{e}}$,
\begin{align}\label{eq:dVde}
\frac{dV^*}{d\mathbf{e}} = \mathbf{e}(t)^\intercal \mathbf{C}(t)^\intercal
\end{align}
where, $\mathbf{C}(t)$ is chosen to be a positive definite matrix, 
then $\dot{\mathcal{V}}(\mathbf{e}(t))$ becomes negative definite.
In order to find the expression for $\mathbf{C}(t)$, we substitute eq. \eqref{eq:dVde} in eq. \eqref{eq:dVde_HJB},
\begin{align}
\mathbf{e}(t)^\intercal \left( \mathbf{I} - \mathbf{C}(t)^\intercal \mathbf{J} \mathbf{R}^{-1} \mathbf{J}^\intercal \mathbf{C}(t) \right) \mathbf{e}(t) = 0
\end{align}
where, $\mathbf{I}$ is $N_p\times N_p$ identity matrix. For this to be true for all $\mathbf{e}(t)$, 
\begin{align}\label{eq:C_HJB}
\mathbf{C}(t)^\intercal \; \mathbf{J} \mathbf{R}^{-1} \mathbf{J}^\intercal \; \mathbf{C}(t) = \mathbf{I}
\end{align}
To find a solution for $\mathbf{C}(t)$, we decompose $\mathbf{J} \mathbf{R}^{-1} \mathbf{J}^\intercal = \mathbf{U}\mathbf{\Sigma} \mathbf{U}^\intercal$ into eigenvectors $\mathbf{U}$ and a diagonal matrix of eigenvalues $\mathbf{\Sigma}$. Since, $\mathbf{J} \mathbf{R}^{-1} \mathbf{J}^\intercal$ is a symmetric positive definite matrix, $\mathbf{U}^\intercal\mathbf{U}^=\mathbf{U}\mathbf{U}^\intercal=\mathbf{I}$ and all eigenvalues are positive. Now, $\mathbf{C}(t)$ can assume the following form to satisfy eq. \eqref{eq:C_HJB}
\begin{align}\label{eq:C}
\mathbf{C}(t) = \mathbf{U} \mathbf{\Sigma}^{-\frac{1}{2}} \mathbf{U}^\intercal
\end{align}
This form assures $\mathbf C(t)$ to be positive definite, so that the system is stable around $\mathbf{e}=0$. However, for numerical stability while implementation, a small positive term ($=10^{-4}\mathbf{I}$) is added to $\Sigma$ so as to avoid numerical instability.

Finally, we get the optimal weight update law by combining equations \eqref{eq:ustar} and \eqref{eq:dVde}  as
\begin{align}
\mathbf{\dot{\widehat{w}}} = \mathbf{u}^*(t) = \mathbf{R}^{-1} \mathbf{J}^\intercal \mathbf{C}(t) \mathbf{e}(t) 
\end{align}
where, $\mathbf{C}(t)$ is given by eq. \eqref{eq:C}.

For the FFNN considered in eqs. \eqref{eq:NN1}-\eqref{eq:NN2}, the Jacobian can be obtained as follows. For the output layer,
\begin{align}
\frac{\partial f(\mathbf{w},\mathbf{x}_p)}{\partial w_{i_2i_1}} = y_{p,i_2}(1-y_{p,i_2})v_{i_1}
\end{align}
and for the hidden layer,
\begin{align}
\frac{\partial f(\mathbf{w},\mathbf{x}_p)}{\partial w_{i_1i_0}} = \sum_{i_2}y_{p,i_2}(1-y_{p,i_2})w_{i_2i_1}v_{i_1}(1-v_{i_1}) x_{p,i_0}
\end{align}

\subsection{Levenberg-Marquardt Modification}

The LM modification is applied to the optimal HJB based learning scheme by changing Eq.~\eqref{eq:C} as
\begin{align}
\mathbf{C}(t) = \mathbf{U} \mathbf{\Sigma}_\mu^{-\frac{1}{2}} \mathbf{U}^\intercal
\end{align}
where, $\mathbf{\Sigma}_\mu$ is a diagonal matrix of the eigenvalues of $\mathbf{J} \mathbf{R}^{-1} \mathbf{J}^\intercal + \mu\mathbf{I}$, and $\mathbf{U}$ consists of columns as the corresponding eigenvectors.
Here, $\mathbf{I}$ is the identity matrix and $\mu$ is the LM parameter. When a weight update $\mathbf{u}$ leads to an increase in error $\mathbf{e}^\intercal\mathbf{e}$, $\mu$ is increased by a fixed factor $\beta$ and a new $\mathbf{u}$ is computed. This step is repeated until the error decreases and the final $\mathbf{u}$ is used to update the FFNN weights. On the other hand, when a $\mathbf{u}$ leads to decrease in error, it is used to update the FFNN weights and $\mu$ is divided by $\beta$.

\section{HJB based Online Learning for FFNN}\label{sec:onlineHJB}
In the online mode, the weight vector is updated after the network is presented with a pattern. In this case, the network dynamics can be presented as
\begin{align}
\mathbf{y} = \mathbf{f}(\mathbf{\widehat{w}})
\end{align}
where the network output is simply observed as a function of network weights ONLY. Here the desired output $\mathbf{y}^d$ is observed and the network response is compared to find 
\begin{align}\label{eq:e_online}
\mathbf{e}=\mathbf{y}^d-\mathbf{y}.
\end{align}
 Thus the network dynamics can be derived following the approach given in Sec.~\ref{sec:offlineHJB}. 

The problem is to find $\mathbf{u}$ so as to minimize
\begin{align} \label{eq:onlineV}
V(\mathbf{e}(t)) = \int_t^\infty \frac{1}{2}(\mathbf{e}^\intercal\mathbf{e} + \mathbf{u}^\intercal \mathbf{R}\mathbf{u}) d\tau
\end{align}
One should note that this global cost function is not pertaining to any specific pattern rather the network is subjected to various inputs while the instantaneous error $\mathbf{e}$ is computed as per Eq.~\eqref{eq:e_online}. 

The optimal instantaneous weight update law, as derived using the HJB equation and Lyapunov stability criterion, is given by
\begin{align}
\mathbf{\dot{\widehat{w}}} = \mathbf{u}^*(t) &= \mathbf{R}^{-1} \mathbf{J}^\intercal \mathbf{C}(t) \mathbf{e}(t)\\
\text{where, } \mathbf{C}(t) =& \mathbf{U} \mathbf{\Sigma}^{-\frac{1}{2}} \mathbf{U}^\intercal
\end{align}
with $\mathbf{U}$ and $\mathbf{\Sigma}$ as the eigenvectors and eigenvalues of $\mathbf{JR}^{-1}\mathbf{J}^\intercal$.

\subsection{Single output Network with Online Learning}\label{sec:singleOut}
Let us consider a special case of single output network
trained in an online fashion. For such a case 
 $\mathbf{J}$ is a $1\times N_w$ matrix, therefore the term $\mathbf{J}\mathbf{J}^\intercal$ is a scalar. 
Assuming $\mathbf{R} = r \mathbf{I}$, the corresponding HJB equation can be written as
\begin{eqnarray}
e^2 - \left(\frac{dV^*}{de} \right) \frac{1}{r} \mathbf{J J}^\intercal \left(\frac{dV^*}{de} \right)^\intercal = 0  \label{eq:sinOut_ambigu} \\
\frac{dV^*}{de } = e \sqrt{\frac{r}{\mathbf{J}\mathbf{J}^\intercal}} \nonumber
\end{eqnarray}
where, $e = y_d - y$.
Therefore, the optimal learning algorithm becomes
\begin{equation}
\mathbf{u^*} = \frac{1}{(r \mathbf{J}\mathbf{J}^\intercal)^{1/2}} \mathbf{J}^\intercal e
\label{eq_02}
\end{equation}

\begin{table} 
  \caption{Average Success Score, in $\%$, for Modulo-2 Function Learning \label{tab:Succ_mod2}}
  \begin{tabular}{|c|ccc|ccc|cc|}\hline
    FFNN & \multicolumn{3}{c|}{Offline} & \multicolumn{3}{c|}{Online} & \multicolumn{2}{c|}{Offline} \\ \cline{2-9}
    architecture & BP & LF & HJB & BP & LF & HJB & LM & HJB-LM \\ [0.5ex] \hline 
    3-4-1 & 24 & 24 & 28 & 0 & 20 & 0 & 84 & 84 \\
    3-6-1 & 80 & 88 & 96 & 36 & 68 & 96 & 96 & 100 \\
    3-8-1 & 84 & 92 & 100 & 56 & 72 & 100 & 100 & 100 \\
    3-10-1 & 90 & 96 & 100 & 80 & 76 & 100 & 100 & 100 \\
    3-15-1 & 100 & 100 & 100 & 100 & 80 & 100 & 100 & 100 \\  [0.5ex] \hline
  \end{tabular}
\end{table}
\begin{table*} 
  \caption{Average number of epochs (with average run time) for successful convergence for Modulo-2 Function Learning \label{tab:Res_mod2}}
  \begin{tabular}{|c|ccc|ccc|cc|}\hline
    FFNN & \multicolumn{3}{c|}{Offline} & \multicolumn{3}{c|}{Online} & \multicolumn{2}{c|}{Offline} \\ \cline{2-9}
    architecture & BP & LF & HJB & BP & LF & HJB & LM & HJB-LM \\ 
    & $\eta=1$ & $\mu=0.05$ & $r=0.1$ & $\eta=1$ & $\mu=0.2$ & $r=5$ & $\mu=10^{-2},\beta=10$ & $\mu=10^{-2},\beta=10$ \\ [0.5ex] \hline 
    3-4-1 & 3047 (7.6 s) & 1087 (9.5 s) & 3519 (6.0 s) & - & 5843 (10.9 s) & - & 266 (7.0 s) & 63 (0.3 s) \\
    3-6-1 & 2880 (4.2 s) & 2801 (9.9 s) & 190 (0.4 s) & 6210 (11.0 s) & 2429 (4.6 s) & 2939 (5.8 s) & 37 (0.15 s) & 25 (0.05 s) \\
    3-8-1 & 1309 (2.4 s) & 480 (10.4 s) & 134 (0.5 s) & 6773 (15.0 s) & 3565 (8.8 s) & 1722 (4.3 s) & 22 (0.06 s) & 26 (0.06 s) \\
    3-10-1 & 1046 (1.4 s) & 447 (0.7 s) & 54 (0.09 s) & 6123 (10.8 s) & 3919 (7.3 s) & 762 (1.5 s) & 18 (0.05 s) & 22 (0.05 s) \\
    3-15-1 & 2544 (3.5 s) & 670 (0.9 s) & 49 (0.07 s) & 3532 (6.4 s) & 4922 (9.5 s) & 731 (1.5 s) & 17 (0.05 s) & 25 (0.05 s) \\  [0.5ex] \hline
  \end{tabular}
\end{table*}


\section{Experiments and Discussion}\label{sec:Results}
The efficiency of the proposed HJB based optimal learning scheme is substantiated with the help of several benchmark learning problems. Notably, the purpose of the presented work is to optimize a given cost function and not explicitly to design robust classification algorithms. Hence, the following experiments aim at optimising the FFNN for a given cost function. Nevertheless, in order to show the generalizability for classification, several experiments have also been included with separate training and testing data over real datasets (Sec. \ref{ssec:BreastCancer}, Sec.~\ref{ssec:CreditApproval}).

The proposed HJB algorithm has been compared with popular learning schemes, viz., BP, LF and LM. BP is the standard back-propagation learning scheme with a constant learning rate $\eta$. LF scheme is a Lyapunov function based learning scheme developed by Behera \emph{et al.}\cite{Behera2006} and has one tunable parameter $\mu$. LM algorithm is an offline algorithm with two tunable parameters - $\mu, \beta$.
In the proposed offline as well as online HJB algorithms, $\mathbf{R}$ has been set to $r\mathbf{I}$, where $\mathbf{I}$ is an identity matrix and $r>0$ is a constant tunable parameter. 

All the algorithms have been implemented in MATLAB R2013a on an Intel(R) Xeon(R) 2.40GHz CPU with 6GB RAM.
Each experiment consists of multiple trials, each starting from a random initial point. For a fair comparison, in a trial, same initial point is chosen for all the learning schemes.

\subsection{Modulo-2 function}
The first example is based on the modulo-2 function, defined in Eq. \eqref{eq:mod2}. 
The FFNN is learned using various algorithms in offline as well as online modes. The simulation is performed for 25 different trials.
The results are tabulated in Tables~\ref{tab:Succ_mod2} and \ref{tab:Res_mod2}, along with the parameter values used.
Table~\ref{tab:Succ_mod2} shows the success score, as defined in eq.~\eqref{eq:successscore}, for each algorithm.
Table~\ref{tab:Res_mod2} presents the average number of epochs as well as average run time for successful convergence of each algorithm.

It can be observed from Table~\ref{tab:Succ_mod2} that, for all these algorithms, achieving 100\% success rate with lesser number of hidden neurons is difficult, but it improves with the increase in the number of hidden neurons. 
In the case of batch (offline) learning, the success score of the proposed HJB algorithm is much higher than both BP and LF algorithms. 
The proposed HJB algorithm achieves more than 95$\%$ success score with 6 or more hidden neurons.
The offline LM algorithm is able to achieve a high success score even with 4 hidden neurons. Nevertheless, the HJB-LM algorithm performs even better. 
It achieves 100\% success score with 6 or more hidden neurons.
On the other hand, learning in the online mode is obviously more difficult, as shown by lower success scores achieved by the online BP, LF and HJB schemes. With 4 hidden neurons, BP and HJB schemes were found unable to converge to global minimum. Still, HJB scheme achieves more than 95\% success score with 6 or more hidden neurons. This implies high probability for convergence to the global minimum.

The corresponding average rates of convergence, can be seen in Table~\ref{tab:Res_mod2}.
In offline mode, the proposed HJB algorithm converges much faster than both BP and LF algorithms.
The number of epochs required for HJB scheme to converge reduces with the increase in the number of hidden neurons, and accordingly, the training time goes down. However, the convergence time improves drastically by introducing the LM modification, as shown by the results of HJB-LM scheme.
On the other hand, for online learning, the number of epochs is higher than that for offline learning. Still, the proposed HJB algorithm converges significantly faster as compared to BP and LF algorithms.

\begin{figure}
  \centering
  \includegraphics[width=\linewidth]{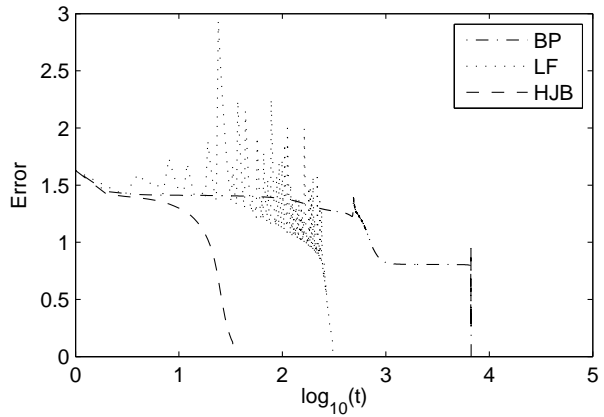}
  \caption{Comparison of convergence rates of diffent algorithms for Modulo-2 function in offline learning.}\label{fig:err_mod2}
\end{figure}

The rate of convergence for one trial each of BP, LF and HJB schemes has been shown in Fig.~\ref{fig:err_mod2}. The error reduces the fastest for the proposed HJB algorithm, followed by LF and BP, respectively. The improvement in the convergence rate with the proposed HJB scheme over that of the other schemes is quite significant, differing by orders of magnitude.

\subsection{8-bit Parity}
The 8 dimensional parity problem is also like the XOR problem where the output is 1 if the number of ones in the input is odd. In this problem, 0 and 1 are replaced by 0.1 and 0.9 respectively. The network architecture has one bias (+1) in the input. 
The simulations are performed for 20 trials with different random initialization points. Table~\ref{parity_tab1} shows the simulation results for different algorithms with 9-30-1 FFNN. 
Both BP and LF based algorithms fail to converge in all the 20 trials. However, the HJB based algorithm converges within 845 epochs with 95\% success score. Interestingly, if one uses HJB LM algorithm, the convergence is achieved within 152 epochs with 100\% success score.
However, it can be seen that LM scheme performs better than the HJB-LM scheme.
Nevertheless, the convergence rate of LM degrades with the increase in the network size. This can be seen in Table~\ref{parity_tab2}, where the HJB-LM scheme converges faster than even LM for a 9-50-1 architecture.
\begin{table}
\caption{8 bit Parity: 9-30-1 architecture} 
\centering                          
\begin{tabular}{c c c c}    
\hline\hline 
Algorithm & Avg Epochs & Success rate   & Parameters\\[0.5ex]
\hline
BP Offline & - & 0\% & $\eta=0.05$  \\
LF Offline & - & 0\% & $\mu = 0.10 $  \\
HJB Offline & 845 & 95\% & $r=5$ \\
LM & 130 & 100\% & $\mu=0.001, \beta = 10$ \\
HJB LM  & 152 & 100 \% & $\mu=0.001, \beta = 10$ \\ [0.5ex]
\hline
\end{tabular}
\label{parity_tab1}

\caption{8 bit Parity Problem : 9-50-1} 
\centering					
\begin{tabular}{c c c c} \hline\hline
Algorithm & Avg Epochs & Success rate   & Parameters\\[0.5ex]
\hline 
HJB Offline & 465 & 100\% & $r=5$ \\
LM  &  129 & 100\% & $\mu=0.001, \beta = 10$ \\
HJB LM  & 97 & 100 \% & $\mu=0.001, \beta = 10$ \\ [0.5ex]
\hline
\end{tabular}
\label{parity_tab2}
\end{table}

\subsection{2D-Gabor Function}
\begin{figure}
\centering
\includegraphics[width=\linewidth]{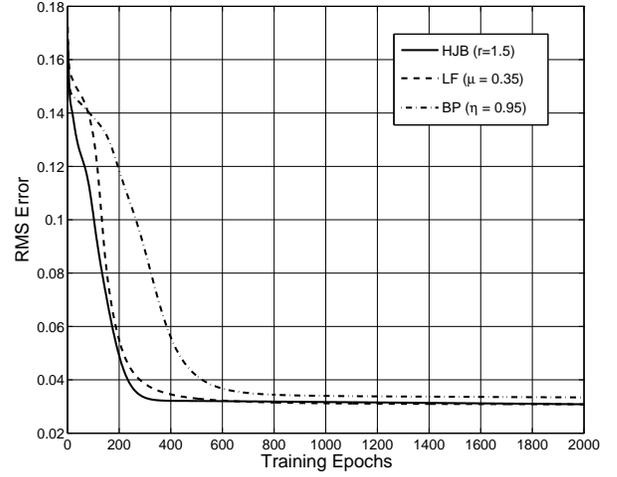}
\caption{Performance comparison between HJB, LF and BP algorithm for 2-D Gabor function}
\label{fig3}
\end{figure}

In  this simulation, Gabor function is used for system identification problem which is represented by the following equation. 
\begin{equation}
\begin{split}
g(x_1,x_2)= \frac{1}{2 \pi (0.5) ^2} &e^{-(x_1^2-x_2^2)/(2 (0.5) ^2)}\\
& \times \text{cos}(2 \pi (x_1+x_2))
\end{split}
\end{equation} 
For this problem, 100 test data are randomly generated in the range [0, 1]. The network architecture used in this problem is 3-6-1, with one bias (+1) input. The simulation is run for online learning till the rms error is 0.01 or a maximum of 2000 epochs. Table~\ref{Tab_gabor} shows the rms error averaged over 20 runs for each method, after 200, 400 and 2000 epochs of learning.
One can observe that the error decay rate is the fastest for the proposed HJB algorithm as compared to that of the other algorithms.
Figure \ref{fig3} compares the error decay for HJB, LF and BP algorithm with epochs.

\begin{table}
\caption{2D Gabor Function: 3-6-1 architecture} 
\centering						  
\begin{tabular}{c c c c c }    
\hline\hline 
Algorithm & \multicolumn{3}{c}{Avg RMS error} & Parameters\\\cline{2-4}
(Online)  & (200 epochs) &  (400 epochs) & (2000 epochs)\\ [0.5ex]
\hline  
BP  & 0.1439 & 0.1404 & 0.0311 & $\eta=0.50$  \\ 
BP  & 0.1222 & 0.0615&  0.0313 & $\eta = 0.95 $ \\ 
LF  & 0.0703 & 0.0354 & 0.0305 & $\mu = 0.35$ \\
HJB  & 0.0488 & 0.0326 & 0.0310 & $r=1.5$ \\ 
HJB  & 0.0479 & 0.0369 & 0.0360 & $r=1$ \\ [0.5ex]
\hline
\end{tabular}
\label{Tab_gabor}
\end{table}

\subsection{Breast Cancer Data} \label{ssec:BreastCancer}
\begin{figure}
\centering
\includegraphics[scale=0.8]{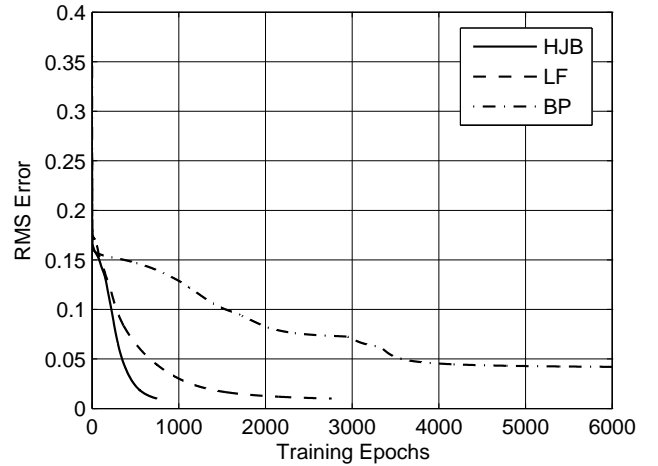}
\caption{Comparison of error decay for HJB, LF and BP algorithm while training for Breast Cancer problem}
\label{cancer_fig1}
\end{figure}

Breast Cancer problem \cite{Bache2013} is a classification problem where the neural networks is fed with 9 different inputs representing different medical attributes while the output represents class of cancer. The data set has 699 data points of which 600 have been used for training and the remaining ones for testing. The misclassification rate is calculated as the percentage of test points wrongly classified by the algorithm. For this problem, a network with 10-15-1 architecture, having one bias (+1) node in the input layer, is trained in an online way. The simulations are performed for 20 different trials. Table~\ref{chap3table2} shows the simulation results. Note that BP algorithm is never converging to 0.01 rms error within 6000 iterations. Fig.~\ref{cancer_fig1} shows the decay of error with successive epochs for HJB, LF and BP algorithms. 



\begin{table}
\caption{Breast Cancer: 10-15-1 architecture} 
\centering						  
\begin{tabular}{c c c c}    
\hline\hline 
Algorithm & Avg. Epochs & Misclassification rate   & Parameters\\[0.5ex]
\hline 
BP Online  & 6000* & 2.95 & $\eta=0.20$  \\ 
LF Online & 3058 & 2.85 & $\mu$ = 0.20  \\ 
HJB Online  & 1020 & 2.05 & $r=2.5$ \\ [0.5ex]
\hline
\end{tabular}
\label{chap3table2}\\
\raggedright{\hspace{2mm} * \footnotesize Error does not fall below the set threshold while training}
\end{table}

\subsection{Credit Approval Data Set} \label{ssec:CreditApproval}
Credit approval \cite{Bache2013} is also classification problem with 14 inputs and a binary output. The output represents the approval or rejection of credit card. The data set has 690 data points of which 80\% i.e. 552 points are used for training and the rest are used for testing. For this problem, we have selected a network with 15-25-1 architecture, with one bias (+1) node in input layer. The training is carried out in online fashion for 20 different trials, starting from different initializations of network weights. Table \ref{Tab_credit} shows the simulation results. Since none of the methods converged to the rms error of 0.01, the training is stopped after 2000 epochs. We have also provided the average rms error after 1000 epochs for each algorithm. We can see that the HJB based algorithm reaches to a better rms error within 1000 epochs as compared to the error for BP and LF based algorithm. Also, note that the BP algorithm does not make much improvement from 1000 to next 2000 epochs. 
\begin{table}
\caption{Credit Approval Data: 15-25-1 architecture} 
\centering						  
\begin{tabular}{c c c c}    
\hline\hline 
Algorithm & Avg RMS  & Avg RMS error & Parameters\\
  & (2000 epochs) &  (1000 epochs) & \\ [0.5ex]
\hline 
BP Online  & 0.1826 & 0.1991 & $\eta=0.20$  \\ 
LF Online  & 0.1379 & 0.1810& $\mu = 0.20 $ \\ 
HJB Online  & 0.0776 & 0.1250 & $r=2.5$ \\ [0.5ex]
\hline
\end{tabular}
\label{Tab_credit}
\end{table}

\subsection{Parameter Tuning}
\begin{figure}
\centering
\includegraphics[width=\linewidth]{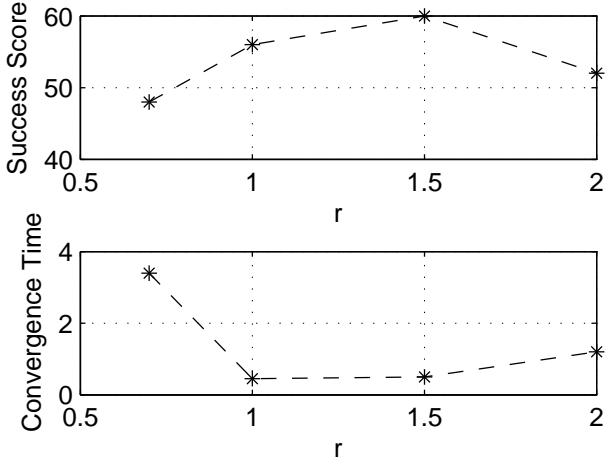}
\caption{Effect of varying the parameter $r$ on the success score and convergence time of HJB offline algorithm for modulo-2 function with 3-4-1 architecture}
\label{fig:paramVar}
\end{figure}
The proposed HJB offline and online algorithms have one tunable parameter, viz., $r$. The value of this parameter plays a crucial role in determining the convergence properties, as shown in Fig.~\ref{fig:paramVar}. 
Notably, BP and LF algorithms also have one parameter each, i.e., $\eta$ and $\mu$, respectively. On the other hand, LM and the proposed HJB-LM algorithms have two parameters each.

Derivation of the optimal values of the parameters has not been dealt with in this work. For the experiments, the parameters have been set manually.

\section{Convergence Behavior}\label{sec:globalMin}
This section analyzes the convergence behavior of various algorithms used in this paper.
The iterative update laws of these algorithms are summarized in Table \ref{tab:allu}.
\begin{table}
  \caption{Weight update laws of various algorithms}\label{tab:allu}
  \centering
  \begin{tabular}{|r|l|} \hline
    Algorithm & Update law, $\mathbf{\dot{\widehat{w}}} = \mathbf{u}$ \\ [0.5ex] \hline
    BP & $\mathbf{u} = \eta \mathbf{J}^\intercal \mathbf{e}$\\ [0.5ex]
    LF & $\mathbf{u} = \mu \frac{||\mathbf{e}||^2}{||\mathbf{J}^\intercal\mathbf{e}||^2} \mathbf{J}^\intercal \mathbf{e}$\\ [0.5ex]
    LM & $\mathbf{u} = [\mathbf{J}^\intercal\mathbf{J}+\mu\mathbf{I}]^{-1} \mathbf{J}^\intercal \mathbf{e}$\\ [0.5ex]
    HJB & $\mathbf{u} = \frac{1}{r} \mathbf{J}^\intercal \mathbf{C} \mathbf{e}$\\ [0.5ex]
    single output online HJB & $\mathbf{u} = \frac{1}{(r \mathbf{J}\mathbf{J}^\intercal)^{1/2}} \mathbf{J}^\intercal e$\\[0.5ex]\hline
  \end{tabular}
\end{table}

\subsection{Global Minimization}
Let us analyze the global convergence properties of the proposed HJB scheme.
The HJB equations for minimizing the cost function of eq.~\eqref{eq:costFnV} provide us with eq.~\eqref{eq:dVde_HJB}, which is underdetermined, i.e. not sufficient to derive an optimal $\mathbf u^*(t)$.
Hence, we take help of Lyapunov function to impose further conditions on the solution. However, it turns out that these conditions compel the error $\mathbf e(t)$ to reduce at each step. This behaves like a greedy search scheme, ending up in the local minimum.
On the contrary, the convergence to global optimum may sometimes require $\mathbf e(t)$ to increase also. 

\begin{figure}
\centering
\includegraphics[scale=0.8]{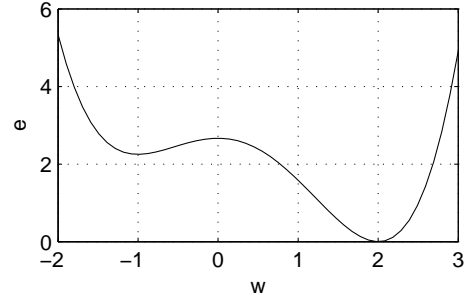}
\caption{An example error function to be minimized with respect to w}
\label{fig:eg_fn}
\end{figure}
This can be analyzed more elaborately with the help of an example.
Consider the following single output function with single parameter, as plotted in Fig.~\ref{fig:eg_fn}
\begin{align}
e = -{f}({w}) = \frac{w^4}{4}-\frac{w^3}{3}-w^2+\frac{8}{3}
\end{align}
While minimizing $e$ with respect to $w$, we use the formulation described in Sec.~\ref{sec:singleOut}. 
Eq.~\eqref{eq:sinOut_ambigu} gives
\begin{align}
{u}^*(t) &= \frac{1}{r} {J} \left(\frac{dV^*}{d{e}}\right) \\
\frac{dV^*}{de } &= \pm e {\frac{\sqrt r}{|J|}} \label{eq:eg_ambigu}
\end{align}
where, $J=df/dw=-de/dw$.
Eq.~\eqref{eq:eg_ambigu} has two solutions, and HJB formulation does not determine which one to choose. 
For global convergence, $w$ should end up in the value 2. 

Consider the case when $w\in(-1,0)$. Here, $dV^*/de < 0$ because if $e$ increases, the global cost function $V^*$ decreases. This makes $u>0$ as $J<0$, taking $w$ to global minimum. 
On the contrary, imposing stability condition with the Lyapunov criterion makes  $dV^*/de>0$, assuming that a decrease in error decreases the global cost function $V$. But this assumption is not true here. This makes $u<0$ as $J<0$, taking $w$ to -1, i.e. a local minimum.
Hence, HJB formulation does not have a provision to choose among the positive or negative form for $dV^*/de$ in eq.~\eqref{eq:eg_ambigu}. Since Lyapunov condition selects always the positive form, HJB with the help of Lyapunov condition cannot guarantee global convergence. Still, the success score is found to increase with HJB as compared to other algorithms. This implies that the local minima are often bypassed due to the greedy nature of the convergence.

\begin{figure}
\begin{center}
  \scalebox{0.5}{\input{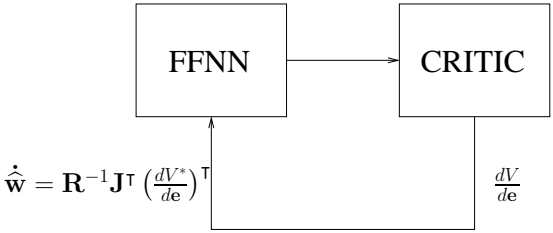}}
  \caption{Single network adaptive critic for estimating the optimal weight update scheme \label{fig:snac}}
\end{center}
\end{figure}

For determining the exact form of $d\mathbf V^*/d\mathbf e$, we have performed some preliminary simulation experiments using a single network adaptive critic \cite{BeheraKar2009,Padhi2006,Govindhasamy2005} (SNAC), as shown in Fig.~\ref{fig:snac}. SNAC makes use of the discrete time HJB equation as explained below.

As in Sec.~\ref{sec:onlineHJB}, the discrete time state dynamics for the online learning of FFNN (refer to eq.~\eqref{eq:FFNNdyn}) can be written as,
\begin{align}
\mathbf{e}_{k+1} = \mathbf{e}_{k}-\mathbf{J}_k \mathbf{u}_k
\end{align}
where, $k$ is the discrete time index and the time step is assumed to be 1 unit. The discrete time global cost function is
\begin{align*}
V_{l} = \sum_{k=l}^\infty \frac{1}{2}(\mathbf e_k^\intercal \mathbf e_k + r \mathbf u_k^\intercal \mathbf u_k)
\end{align*}
The optimal cost-to-go function is,
\begin{align*}
V^*_{k} = \min_{\mathbf u}\{\frac{1}{2}(\mathbf e_k^\intercal \mathbf e_k + \mathbf u_k^\intercal \mathbf u_k) + V^*_{k+1} \}
\end{align*}
which, on differentiation with respect to $\mathbf u_k$ and $\mathbf e_k$, gives
\begin{align}
\mathbf{u}_k = \frac{1}{r} \mathbf{J}_k^\intercal \mathbf \lambda_{k+1}\\
\mathbf\lambda_k = \mathbf e_k + \mathbf\lambda_{k+1} \label{eq:lambda}
\end{align}
where, $\mathbf\lambda_k=dV_k/d\mathbf e_k$. A simple 2-layer FFNN, called Critic network, is used to estimate $\mathbf\lambda_{k+1}$ as a function of $(x_k)$, subject to eq.~\eqref{eq:lambda}. The implementation details can be found in Behera and Kar \cite{BeheraKar2009}. 

The SNAC based scheme has been implemented for modulo-2 function in online mode with a 3-4-1 (+1 bias in input) FFNN, with a 2-6-1 (+1 bias in input) FFNN as critic network. The scheme shows a success score of 80\% over 20 trials. This is quite large as compared to other online algorithms shown in Table~\ref{tab:Succ_mod2}, where online HJB scheme has 0\% success score and batch mode HJB has only 28\% success score. 
However, these results are very rudimentary.
Nevertheless, they show that a properly tuned adaptive critic can possibly achieve 100\% success score. This investigation forms the future scope of this paper.

Now let us analyze the global convergence behavior of other algorithms.
Any deterministic weight update law considered in this paper should take the weights to local minimum if the corresponding Lyapunov function is negative semi-definite. The Lyapunov function is given by
$\mathcal{V}(\mathbf{e}) = \frac{1}{2}\mathbf{e}^\intercal \mathbf{e}$.
The equilibrium point $\mathbf e=0$ is asymptotically stable if $\dot{\mathcal{V}}(\mathbf{e})$ is strictly negative definite.
For any deterministic weight update scheme for a general FFNN, $\mathbf{\dot{e}}$ can be expressed as in Eq.~\eqref{eq:edot}. Thus,
\begin{align}
\dot{\mathcal{V}}(\mathbf{e}) = -\mathbf{e}^\intercal \mathbf{J} \mathbf{u}(t)
\end{align}
For HJB scheme, 
\begin{align}
\dot{\mathcal{V}}(\mathbf{e}) = -\frac{1}{r} \mathbf{e}^\intercal \mathbf{J} \mathbf{J}^\intercal \mathbf{C} \mathbf{e}
\end{align}
is negative semi-definite as $\mathbf{J} \mathbf{J}^\intercal$ can be positive semi-definite.
For LM scheme also
\begin{align}
\dot{\mathcal{V}}(\mathbf{e}) = - \mathbf{e}^\intercal \mathbf{J} [\mathbf{J}^\intercal\mathbf{J}+\mu\mathbf{I}]^{-1} \mathbf{J}^\intercal \mathbf{e}
\end{align}
is negative semi-definite, as $\mathbf{J}=0$ at local minima. 
Moreover, it has already been pointed out in Behera {\sl et al.} \cite{Behera2006} that the LF algorithm leads to a negative semi-definite Lyapunov function.
This implies that the above algorithms cannot guarantee global convergence.

\subsection{Computation time and complexity}
It has been observed that the proposed HJB algorithm is much faster than the other algorithms for both online as well as offline learning.
It will be useful at this stage to compare the computational complexity of one iteration of each of these algorithms.
All the algorithms discussed in this paper are based on deterministic iterative update of weights. Moreover, all of these involve the calculation of the Jacobian $\mathbf{J}$. They mostly differ in their weight update schemes, $\mathbf{u}$.

Certainly, a single iteration of HJB and LM schemes involves more computations than BP and LF schemes. However, the number of iterations required for HJB and LM are significantly small as compared to those for BP and LM. Hence, the former ones take much less convergence time than the later ones.

The Jacobian $\mathbf{J}$ is an $N_oN_p\times N_w$ matrix.
HJB involves calculation of the eigenvalues of $\mathbf{J}\mathbf{J}^\intercal$ ($N_oN_p\times N_oN_p$), while LM involves calculation of the inverse of $\mathbf{J}^\intercal\mathbf{J}$ ($N_w\times N_w$).
Hence, in the offline mode, where $N_p$ is large, one iteration of LM is supposed to take lesser time than HJB.
Nevertheless, the number of iterations required for HJB is lesser than that for LM, leading to reduction in the computation time for the former. Moreover, as the size of network, $N_w$, increases, the performance of LM deteriorates quickly; but since $\mathbf{J}\mathbf{J}^\intercal$ is independent of $N_w$, HJB is not affected drastically.
On the other hand, in online mode, $N_p=1$, and hence, HJB scheme becomes even faster.

\section{Conclusion} \label{sec:Concl}
HJB equation is well known for dynamic optimization. In this paper, the weight update process in FFNN has been formulated as a dynamic optimization problem. Applying HJB equations, the optimal weight update laws have been derived in both batch and online (instantaneous) modes.
However, the HJB equation turns out to be under-determined, i.e. $dV/d\mathbf e$ has multiple solutions, w.r.t. eq.~\eqref{eq:dVde_HJB} and it turns out that we have to impose additional conditions to ensure global minimization. A proper choice of this solution also requires the direction where the global minimum is located. This issue requires further investigation.
This paper analytically analyzes the convergence behavior of various popular learning algorithms as why they are liable to be stuck in local minima.
It is shown that the Lyapunov function drags the solution to the local minimum.
We have performed an initial simulation experiment using single network adaptive critic \cite{BeheraKar2009} to find the global convergence for modulo-2 function. The results are very  encouraging and will form the basis of our further investigation.

Nevertheless, although the global optimization issue remains unresolved, the proposed scheme ensures faster convergence rate as compared to the existing schemes, including LM. Moreover, the weight update law using HJB equation has been derived for both offline as well as online modes, whereas LM can be applied only in the offline mode. Our study also shows that the proposed HJB scheme works well even with large network size, while the LM method deteriorates in performance as the size of the network increases, as also observed by Xie {\sl et al} \cite{Xie2012}.

\bibliographystyle{IEEEbib}
\bibliography{HJB}

\end{document}